# A deep Convolutional Neural Network for topology optimization with strong generalization ability

Yiquan Zhang[a] · Bo Peng[a] · Xiaoyi Zhou[a] · Cheng Xiang[a] · Dalei Wang[a*]

**Abstract** This paper proposes a deep Convolutional Neural Network(CNN) with strong generalization ability for structural topology optimization. The architecture of the neural network is made up of encoding and decoding parts, which provide down- and up-sampling operations. In addition, a popular technique, namely U-Net, was adopted to improve the performance of the proposed neural network. The input of the neural network is a well-designed tensor where each channel includes different information for the problem, and the output is the layout of the optimal structure. To train the neural network, a large dataset is generated by a conventional topology optimization approach, i.e. SIMP. The performance of the proposed method was evaluated by comparing its efficiency and accuracy with SIMP on a series of typical optimization problems. Results show that a significant reduction in computation cost was achieved with little sacrifice on the performance of design solutions. Furthermore, the proposed method can intelligently give a less accurate solution to a problem with the boundary condition not included in the training dataset.

# 1 Introduction

Since the seminal paper of Bendsoe and Kikuchi (1988), studies on structural topology optimization problems, which can be traced back to Michell (1904), have made tremendous progress. A variety of numerical methods have sprung up later, including SIMP (Bendsoe, 1989; Zhou and Rozvany, 1991; Rozvany et al., 1992), evolutionary approaches (Xie and Steven, 1993), moving morphable components (Guo et al., 2014; Zhang et al., 2018), level-set method (Wang et al., 2003; Allaire et al., 2004; Du et al., 2018), and others. However, the computational cost is still one of the main obstacles preventing the adoption of topology optimization methods in practice, in particular for large structures (Sigmund and Maute, 2013).

With the recent boost of machine learning algorithms and advances in graphics processing units (GPU), machine learning (ML), especially the deep learning, has been seen to make many successful stories in various fields, including automatic drive, image recognition, natural language processing, and even art. It may shed light on accelerating the adoption of topology optimization in more design practices. Recently, a few attempts have been seen to apply ML on topology optimizations (Lei et al., 2018; Sosnovik and Oseledets, 2017; Banga et al., 2018; Yu et al., 2018), microstructural materials design (Yang et al., 2018) and additive manufacturing (Nagarajan et al., 2018). Theoretically, the optimal layout of the material is a complicated function of the initial conditions based on the optimization objective and constraints. The neural network is good at fitting a complicated function and this characteristic makes it possible for the neural network to fit a target function which can directly give us a good structure without any iteration and effectively reduce computational time.

Sosnovik and Oseledets (2017) first introduced the deep learning model to topology optimization and improved the efficiency of the optimization process by formulating the problem as an image segmentation task.

Dalei Wang
E-mail: wangdalei@tongji.edu.cn
* Corresponding Author
[a] Department of Bridge Engineering, Tongji University, 1239 Siping Road, 200092, Shanghai, China.

His deep neural network model could map from the intermediate result of the SIMP method to the final structure of the design, which effectively decreased the total time consumption. However, his work did not consider the initial conditions for topology optimization, and the accuracy of the result heavily relies on the first few iterations. Banga et al. (2018) proposed a deep learning approach based on a 3D encoder-decoder Convolutional Neural Network architecture for accelerating 3D topology optimization and to determine the optimal computational strategy for its deployment. Elaborating as Sosnovik and Oseledets (2017), this method also takes the intermediate result of the conventional method as the input of the neural network. It could cut down some time needed because of the reduction of iterations, but it can not completely replace the traditional method.

Lei et al. (2018) developed a ML driven real-time topology optimization paradigm under the Moving Morphable Component-based solution framework. Their approach can reduce the dimension of parameter space and enhance the efficiency of the ML process substantially. The ML models used in the approach were the supported vector regression and the K-nearest-neighbors. Rawat and Shen (2018) proposed a new topology design procedure to generate good-performance structures using an integrated Generative Adversarial Network (GAN) and CNN architecture. But only volume fraction, penalty and radius of the filter are changeable in this method. All other initial conditions including force and boundary conditions must be fixed. Guo et al. (2018) proposed an indirect low-dimension design representation to enhance topology optimization capabilities, which can simultaneously improve computational efficiency and solution quality. Yu et al. (2018) also used the GAN and CNN to propose a deep learning-based method, which can predict an optimized structure for a given boundary condition and optimization setting without any iteration. However, the neural network model trained by a large dataset in his research could work under only one boundary condition. Therefore, it is impractical to work in the real world because the time needed for preparing the dataset and training the model is much longer than the traditional SIMP method. To reduce time on preparing data for training, Cang et al. (2019) proposed a theory-driven learning mechanism which uses domain-specific theories to guide the learning, thus distinguishing itself from purely data-driven supervised learning. Oh et al. (2019) proposes an artificial intelligent (AI)-based deep generative design framework that is capable of generating numerous design options which are not only aesthetic but also optimized for engineering performance.

In this study, a deep CNN model with strong generalization ability is proposed to solve topology optimization problems. The input of our network is a multi-channel array with each channel representing different initial conditions and the output is a segmentation mask with each element represents the probability of reservation. The evaluation result shows that the proposed method can predict a near-optimal structure in negligible time. The main novelty of this work is the strong generalization ability of the proposed deep CNN. It can give a less accurate solution to the topology optimization problems with the different boundary conditions even though the CNN was trained on one boundary condition. The proposed method will be useful in the preliminary design stage as the well-trained neural network is able to deliver a rough result which provides the designers with a general idea within a very short time.

## 2 Overview of neural networks

2.1 Artificial neural networks and convolutional neural networks

Artificial neural networks (ANN), inspired by animal nervous systems, is one of the most prevalent and successful algorithms in machine learning. The basic component of the neural network is a neuron, which is a mathematical approximation of real neurons. In neural systems, a neuron could accept signals from other neurons, and if the aggregated signals exceed the threshold, the neuron is fired and then sends a signal to the related ones. In ANNs, a neuron is an abstract computation unit, receiving inputs from the former neurons and returning an output to other neurons, which can be mathematically expressed as:

$$\boldsymbol{y} = f(\boldsymbol{z}) = f(\boldsymbol{w}^T\boldsymbol{x} + \boldsymbol{b}) \qquad (1)$$

where $\boldsymbol{x}$ is the $n$ dimensional input vector, $\boldsymbol{w}$ is the $n$ dimensional weight factor vector, and $b$ is the bias vector. The reason why ANNs have powerful fitting capabilities lies in $f(\boldsymbol{z})$, which is known as an activation function and is usually a non-linear function. There are several commonly used activation functions, including the sigmoid function, the hyperbolic tangent activation function $tanh(x)$ and the ReLU function (Nair and Hinton, 2010). They map a linear input $\boldsymbol{w}^T\boldsymbol{x} + \boldsymbol{b}$ to a non-linear form to strengthen the expression capabilities of a network.

The prevalence of neural networks in recent years mainly stems from the wide application of convolutional neural networks. Early in the nineties of the last century, Lecun et al. (1998) introduced the convolution operation to ANNs for handwritten numeral recognition, and with the growth of data and computation capabilities, CNNs are widely applied in computer vision, natural language processing and other relevant fields. In CNNs the inner product part $w^T x + b$ is replaced by convolution operation and for 2-D images convolution is defined as:

$$w(x,y) * f(x,y) = \sum_{s=-a}^{a}\sum_{t=-b}^{b} w(s,t) f(x+s, y+t) \quad (2)$$

where $*$ denotes the convolution operation; $f(x,y)$ denotes the input image; $w(x,y)$ denotes the convolution kernel.

A modern CNN architecture is a stack of layers including convolution layers for convolution operation, pooling layers for dimension reduction and fully connection layers (or dense layers) which work like common ANNs. Some advanced CNNs have more complicated topology network architecture for different tasks. For example, GoogLeNet (Szegedy et al., 2014; Ioffe and Szegedy, 2015; Szegedy et al., 2016; Szegedy et al., 2016) introduced "inception module" which combines convolution with various kernels for feature infusion. ResNet (He et al., 2016) proposed skip-connection for training very deep CNNs.

2.2 Semantic segmentation

The semantic segmentation in computer vision is the process of classifying each pixel to a specific category in a given image. Take the self-driven car as an example: each pixel in the image from the front cameras is classified as the road, pedestrian or other vehicles by the neural network so that the self-driven cars could understand the scene in front of themselves. The goal of the topology optimization is to determine whether each element of a structure should be retained or discarded after the optimization algorithm, which can be taken as classifying pixels into two categories. In this sense, topology optimization is equivalent to the semantic segmentation.

Semantic segmentation has realized great achievements with the help of CNNs in recent years. Traditional semantic segmentation algorithms highly rely on hand-design features for the input image, such as color or texture distribution. But in the CNN based methods, these features can be learned automatically through convolution operation so that the segmentation process is greatly simplified. In this sense, each convolution operation can be deemed as a feature extractor which could discover proper features and reduce man-made recognition bias for the segmentation tasks. Shelhamer et al. (2014) firstly introduced CNN in the field of semantic segmentation and named their segmentation network as fully convolutional networks (FCN) since they discarded all the dense layers in their network. Badrinarayanan et al. (2017) trained a neural network based on the encoder-decoder architecture and proposed an "unpooling" operation for upsampling low-resolution images. To improve the performance of encoder-decoder architecture, Ronneberger et al. (2015) proposed a U-shape architecture to implement feature map fusion with shallow and deep layers and named the proposed architecture as U-Net. The key of U-Net lies in the feature fusion. As the network becomes deeper, the feature map encodes more semantic information but loses spatial information since the resolution is reduced. So it is hard for traditional CNNs to strike a balance between semantic and spatial information and achieve a good performance in the segmentation task. While in the U-Net architecture, the shallow layers with spatial information are combined with deeper layers which encode more semantic information, so that the network could utilize the information in all layers and the performance of the network is improved.

3 The typical topology optimization problem

In this study, the proposed method is described alongside with the classical compliance minimization problem solved by SIMP, while other optimization problems are also fit to the framework in principle. The compliance minimization problem is formulated as follows:

$$\min_{x} : \quad c(\mathbf{x}) = \mathbf{U}^T \mathbf{K} \mathbf{U} = \sum_{e=1}^{N} E_e(x_e) \mathbf{u}_e^T \mathbf{k}_0 \mathbf{u}_e \quad (3)$$

$$\text{subjected to:} \quad \frac{V(\mathbf{x})}{V_0} = f, \quad (4)$$

$$\mathbf{K}\mathbf{U} = \mathbf{F}, \quad (5)$$

$$0 \leq \mathbf{x}_{min} \leq \mathbf{x} \leq 1 \quad (6)$$

where $c$ is the compliance, $\mathbf{K}$ is the global stiffness matrix, $\mathbf{U}$ and $\mathbf{F}$ are the displacement and force vectors, respectively, $\mathbf{u}_e$ is the element displacement vector, $\mathbf{k}_0$ is the element stiffness matrix for an element with fully distributed solid material, $\mathbf{x}$ is the vector of design variables (i.e. the element relative densities), $\mathbf{x}_{min}$ is the lower bound, which aims to avoid singularity, $N$ is the number of elements used to discretize the design domain, $V(x)$ and $V_0$ are the material volume and design domain volume, respectively, and $f$ is the prescribed volume fraction.

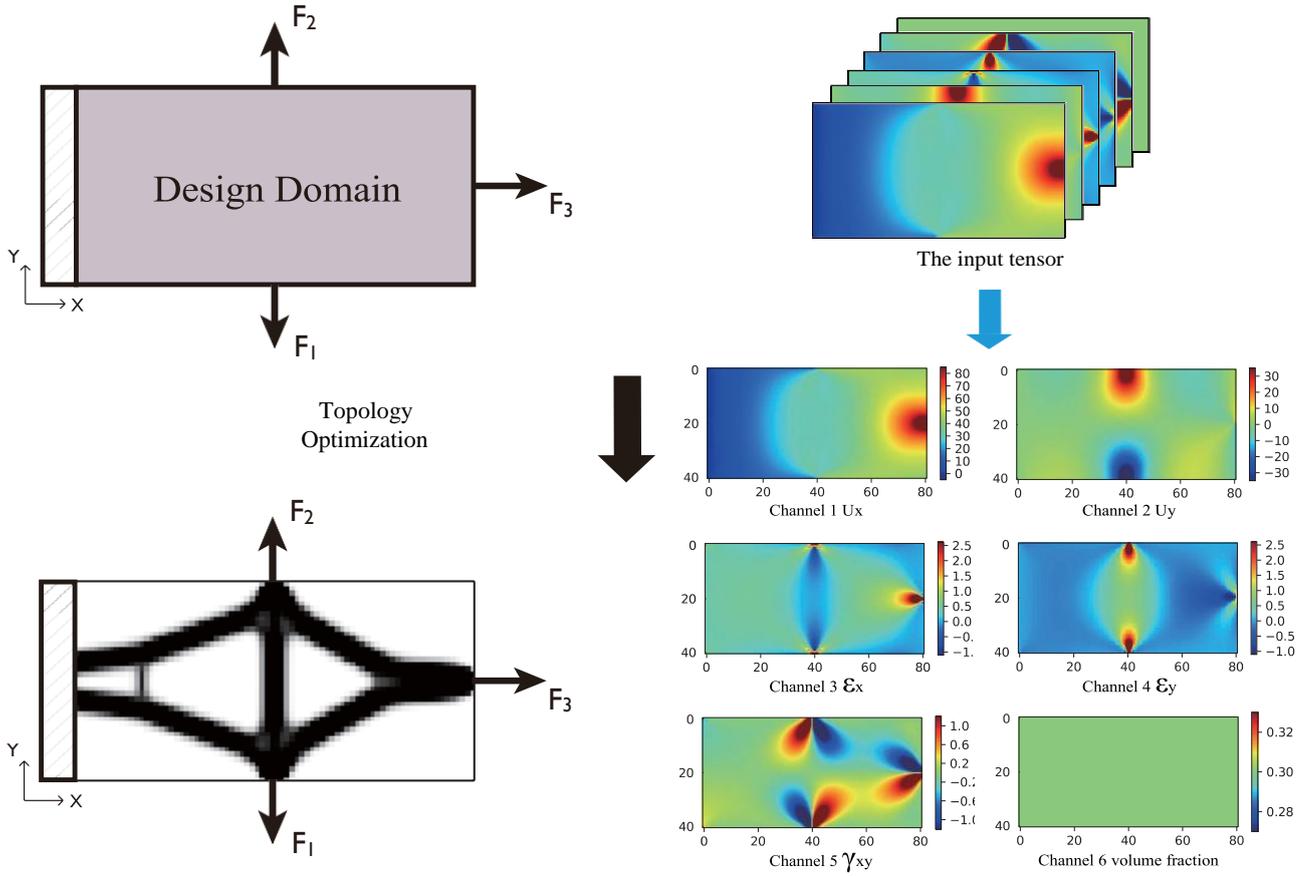

**Fig. 1** The input and output of neural network

## 4 The proposed deep CNN

### 4.1 The input and output of the neural network

For preparing the input of the neural network, a dataset of 80000 samples was generated with the 88 lines of code (Andreassen et al., 2011) which uses the SIMP method. It is worth mentioning that it is actually not necessary to use the SIMP method, other structural topology optimization methods can serve the same purpose. The volume fraction, number of forces and direction of each force are three uniform distributed variables in the initial conditions. The details of the samples are listed as follows:

  1) Resolution: 40 × 80
  2) Cantilever beam boundary condition (Fig.1)
  3) Volume fraction: 0.2 - 0.8 (Uniform distribution)
  4) Penalty factor: 3
  5) Filtering radius: 1.5
  6) Number of forces: 1 - 10 (Uniform distribution)
  7) Direction of each force: $X^+$, $X^-$, $Y^+$, $Y^-$ (Uniform distribution)

The input of the neural network is a 41 × 81 × 6 tensor which includes the initial displacement field, strain field and volume fraction. Therefore, nodal displacements and strains must be calculated by finite element method under the assumption that all elements' relative densities are equal to the volume fraction. A similar idea was proposed in Gao et al. (2017)'s paper where authors used the initial stress field to predict a good ground structure. Fig.1 shows an example of the input with the volume fraction equal to 0.3. The first two channels of the input tensor represent the nodal displacements in $X$ and $Y$ directions, respectively. The next three channels are made up of the nodal normal strains $\varepsilon_x$, $\varepsilon_y$ and shearing strain $\gamma_{xy}$. All the numbers of the last channel are identical to the volume fraction.

The output of the MATLAB code (Andreassen et al., 2011) gives the layout of the optimized structure under the given condition in the form of relative density, which is a 40 × 80 matrix with elements ranging from 0 to

1. It can be taken as a probability matrix representing the probability that a pixel should be retained. Hence, the optimal design can be directly treated as the goal output of the neural network.

To efficiently train the neural network and accurately evaluate its performance, all the samples are divided into the training set, validation set and test set with a ratio of 8:1:1.

4.2 The architecture of the neural network

Our network could be divided into two parts:

1) Encoding part: down-sampling the given array and return a dimension reduced array;

2) Decoding part: up-sampling the given array and return a dimension ascended array.

Fig. 2 shows the architecture of the neural network. Considering the fact that the input of our network is the node condition and the output is the element solution, the input tensor needs to be modified before sent to encoding blocks. The input tensor with 6 channels is firstly sent to a convolution layer with no padding to reduce the shape. The next 3 blocks are encoding blocks with each block consisting of 2 convolution layers, 2 batch norm layers, and 1 pooling layer, so an input tensor is convoluted, normalized and pooled in each encoding block. The final output of encoding part is a multi-channel tensor with 8 times reduced in shape compared to the original.

Before sent to decoding blocks, the outputs of the encoding blocks are sent to 2 additional convolution layers to generate feature maps. Next, the feature maps are sent to decoding blocks, each of which consists of concatenate layer, transpose-convolution layer, batchnorm layer, and convolution layer. There are two kinds of inputs for each decoding block: one is from the former convolution layer and the other is from the corresponding encoding block. The inputs are first concatenated in the channel dimension. Then in the transpose-convolution layer, the dimension reduced array is up-sampled to restore the shape. And a convolution layer is followed to generate the shape restored feature map.

So after 3 decoding blocks, we could get an array with the same shape as the input of the encoding part.

In the last layers of our network, we use a stack of convolution layers with batch normalization layer to obtain element solution. The outputs of the decoding part are firstly concatenated with the input of encoding part in channel dimension. Then the channels are reduced in the following convolution layers and a feature map with only one channel is generated in the final layer. To obtain the reservation probability for each element, the activation of this final layer is set as the sigmoid function.

The model was trained on a computer with an Intel(R)Core(TM) i7-8700k CPU and an NVIDIA GeForce GTX1080 Ti GPU using the Keras framework. The l2 regularizer weight is 1e-5 and the initial learning rate is 0.001. When the validation loss does not decrease for 10 epochs, the learning rate will be decayed by 0.1 times. The details about the loss function, the regularization method and the optimizer are shown in 4.2.1 and 4.2.2.

*4.2.1 Loss function and regularization*

In the view of optimization, the training process of a neural network is equivalent to an optimization problem where we need to minimize or maximize an objective function of the input tensor and other parameters. In this paper, the objective of topology optimization is deemed as searching for a probability distribution since each element's relative density is normalized to $(0, 1)$ denoting the probability of existence. So given the topology optimization target, we need to minimize the "distance" between the known distribution (optimiza- tion target) and the output distribution, and Kullback- Leible divergence is proposed:

$$D_{KL}(p||q) = \sum_x p(x) \log \frac{p(x)}{q(x)} \qquad (7)$$

where $p(x)$ is the optimization target (or empirical distribution); $q(x)$ is the output distribution of neural network. Kullback-Leible divergence is widely used in machine learning, and it has many equivalent variants such as cross entropy, ordinary least squares, etc.

Besides, regularization is also used in designing loss function. Regularization is a penalty added on the parameters to prevent overfitting and L2 regularization (also known as L2 norm) is widely used:

$$\Omega(\theta) = \frac{1}{2} \sum_i \theta_i^2 \qquad (8)$$

where $\theta_i$ is the network parameters such as convolution kernels.

Therefore, the total loss function is the sum of Kullback-Leible divergence and L2 regularization:

$$L = D_{KL}(p||q) + \lambda \Omega(\theta) \qquad (9)$$

where $\lambda$ is the regularization weight, which denotes the importance of regularization compared to Kullback-Leible divergence.

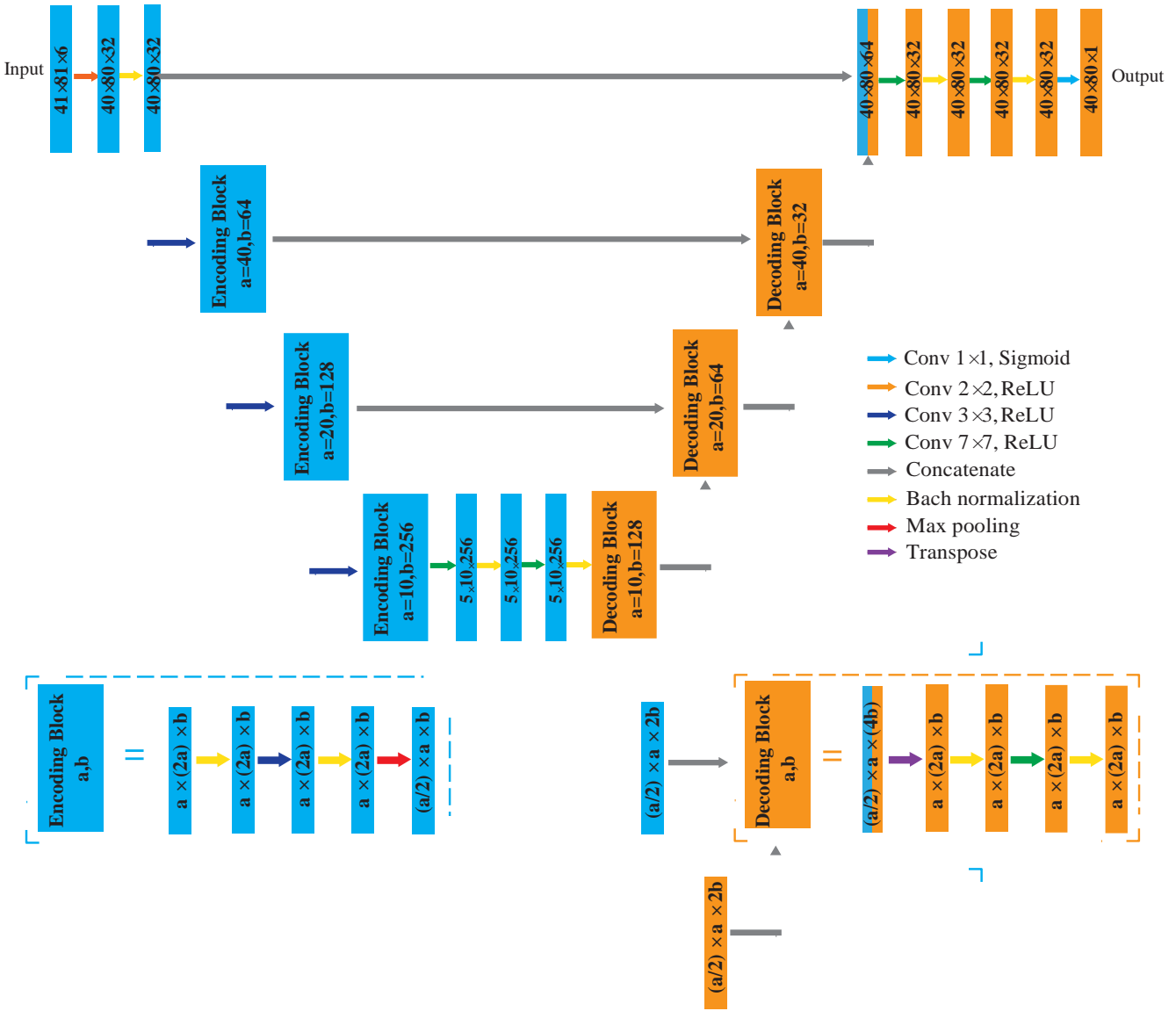

**Fig. 2** The architecture of the neural network

*4.2.2 Optimizer*

In neural networks, gradient based methods are widely used to optimize the loss function and the common form is:

$$\theta_t = \theta_{t-1} - a \frac{\partial L}{\partial \theta} \tag{10}$$

where $\theta_t$ is the parameter $\theta$ in $t$ step; $a$ is the learning rate and $L$ is the loss function.

One of the drawbacks of vanilla gradient descent methods is that the choice of learning rate $a$ may have a great effect on the training process. In most cases, a smaller learning rate may delay the training process and a larger learning rate could even lead to the failure of training. Therefore, some advanced optimization algorithms are proposed to accelerate the training process of modern neural networks. In this paper, the Adam algorithm (Kingma and Ba, 2014), which adaptively modifies the actual learning rate with the first and second order moment estimation of the gradient for each parameter, is used and the updating rule is:

$$g_t = \frac{\partial L}{\partial \theta} \tag{11}$$

$$m_t = \beta_1 \cdot m_{t-1} + (1 - \beta_1) \cdot g_t \tag{12}$$

$$v_t = \beta_2 \cdot v_{t-1} + (1 - \beta_2) \cdot g_t^2 \tag{13}$$

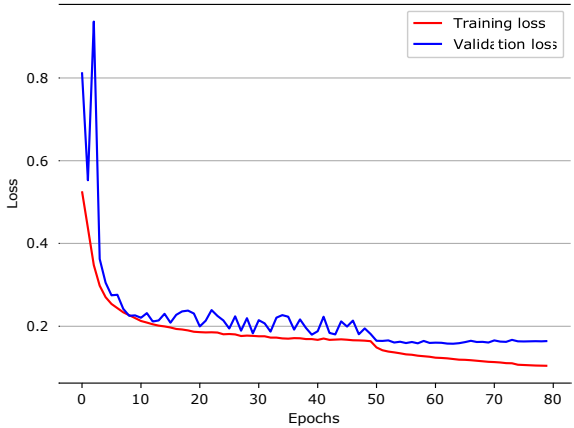

**Fig. 3** The loss on training and validation set

$$\hat{m}_t = \frac{m_t}{1 - \beta_1^t} \qquad (14)$$

$$\hat{v}_t = \frac{v_t}{1 - \beta_2^t} \qquad (15)$$

$$\theta_t = \theta_{t-1} - \alpha \cdot \frac{\hat{m}_t}{\sqrt{\hat{v}_t} + \varepsilon} \qquad (16)$$

where $m_t$ is the biased first moment estimate of gradients; $\beta_1$ is the exponential decay rate of $m_t$; $v_t$ is the biased second moment estimate of gradients; $\beta_2$ is the exponential decay rate of $v_t$; $\hat{m}_t$ is the bias-corrected first moment estimate of gradient; $\hat{v}_t$ is the bias-corrected second moment estimate of gradient.

Fig.3 shows the history of loss on training and validation dataset during the training process. The validation loss converges to about 0.15 after 50 epochs while the training loss still goes down and the overfitting appears. Since the validation loss is the only indicator we focus on, more epochs of training are unnecessary. The gap between validation loss and training loss can be reduced by adjusting the hyperparameter $\lambda$ in equation (9), which is meaningless because it will increase the validation loss.

## 5 Performance evaluation of the method

The performance of the deep neural network model was evaluated by the 8000 samples in the test set. The structure designs of these samples were calculated by both the conventional and proposed method. All the calculations were performed on a computer with an Intel(R) Core(TM) i7-8700k CPU and an NVIDIA GeForceGTX 1080 Ti GPU. The average computation time with the conventional method is 3.913s, while the proposed takes 0.001s in average. It should be noted that the considerable time of generating the dataset and training the neural network are not included, because once the neural network has been well trained, it can be used to solve innumerable topology optimization problems and these times are negligible as the number of the solved problems increases. In this sense, the computational efficiency is greatly improved.

In most cases, there are slight differences between the final structures calculated by the conventional method and the proposed method, as shown in Fig.4(a). The details of the result comparison are listed in Table 1. Nevertheless, 4.12% samples in the test set have huge compliance errors due to the emergence of the structural disconnection (Fig.4(b)). A similar observation was also reported in the Yu et al. (2018)'s work. It is a shortcoming for this kind of method since disconnections in material are completely unacceptable for structures. In the proposed method, the high pixel similarity is the only training objective of the deep neural network, which means the appearance of structural disconnection is not detested as long as the result has high pixel accuracy. Therefore, the structure compliance should be included in the neural network's training objective in future work. Besides the unacceptable results, about 8% results are a pleasant surprise. The structure predicted by the neural network has lower compliance with smaller volume fraction than the structure calculated by the conventional method, as shown in Fig.4(c). They look very similar to each other, but the boundary of the predicted structure is clearer. In other words, the pixel values in the predicted structure are closer to 0 or 1. Although the surprising results are only a small fraction of the test set, their appearance gives the hope that the proposed method may perform better in both computational efficiency and optimality than conventional method after the continuous improvement in the future.

**Table 1** The result of comparison

|  | Compliance error | Pixel values error | Volume fraction error | Calculation time saved |
|---|---|---|---|---|
| Average value [1] | 4.16% | 4.53% | 0.13% | 99.97% |

[1] Samples with structural disconnection are not included.

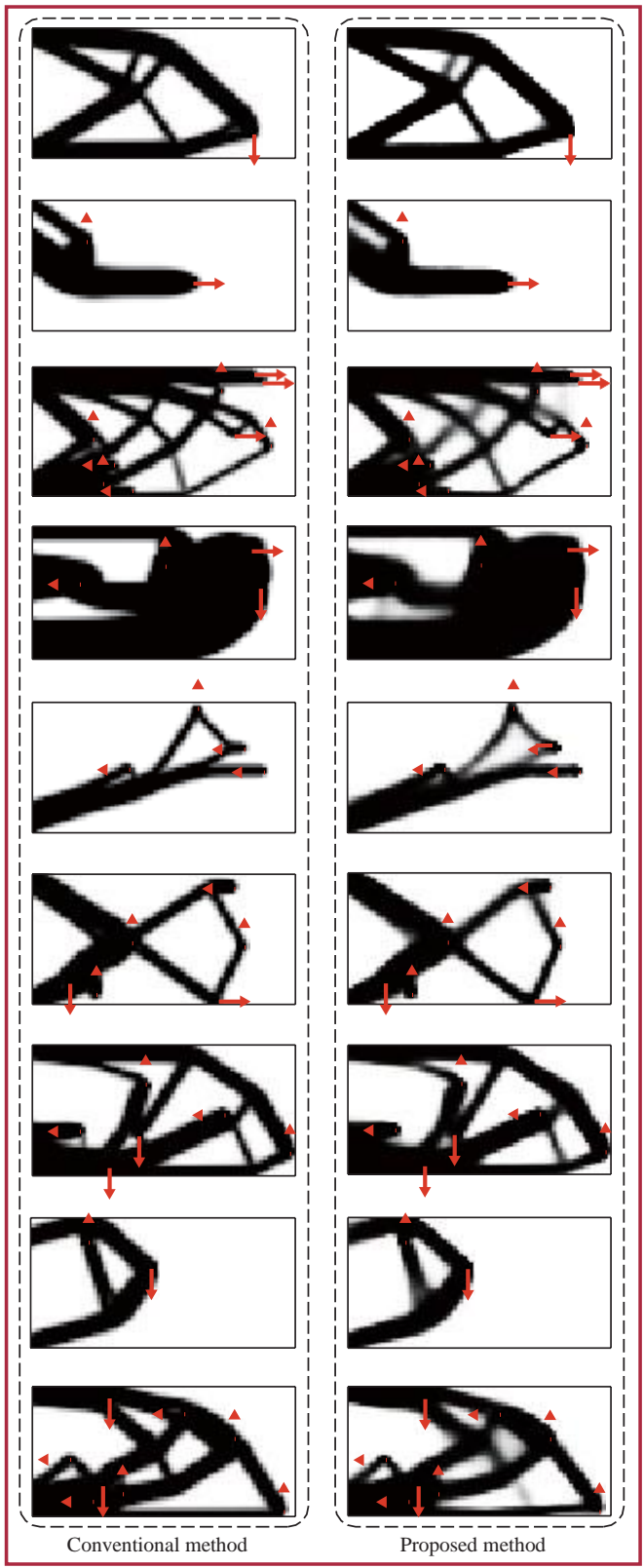

(a) Examples of the normal case

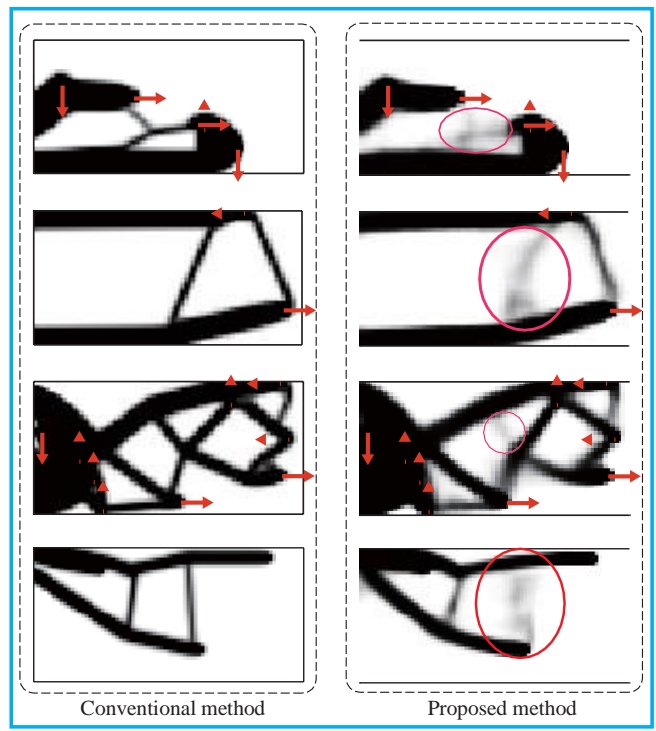

(b) Examples of emergency of structural disconnection

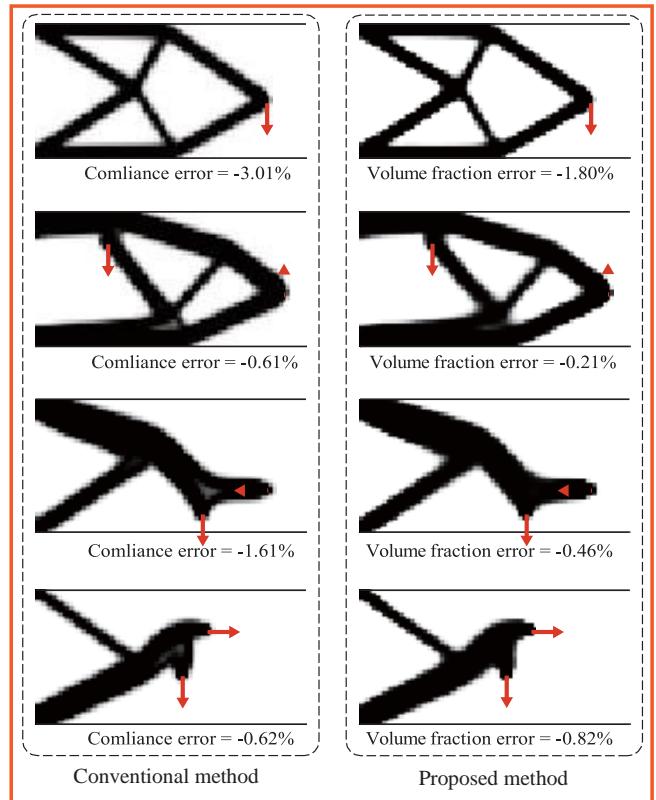

(c) Examples of the results better than expectation

**Fig. 4** Comparison of conventional and proposed method

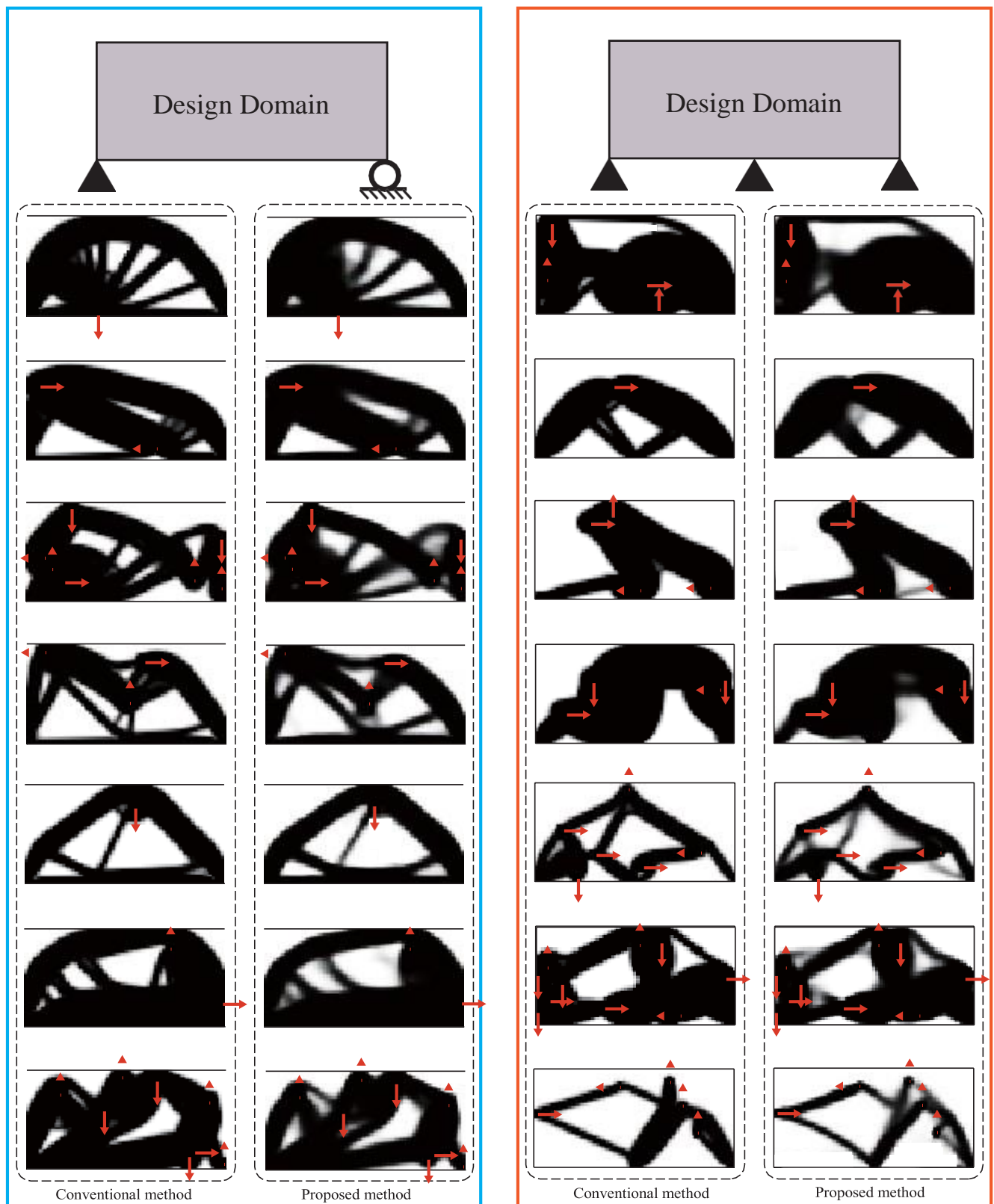

(a) Simply supported beam boundary condition     (b) Continuous beam boundary condition

**Fig. 5** Examples of the results on other boundary condition

**Table 2** The performance of the proposed neural network on other boundary condition

| Boundary condition | Compliance[1] error | Pixel values[1] error | Volume fraction[1] error | Percentage of structural disconnection |
|---|---|---|---|---|
| Simply supported beam | 7.63% | 8.05% | 1.22% | 13.64% |
| Continuous beam | 7.14% | 7.47% | 0.93% | 11.50% |

[1] Samples with structural disconnection are not included.

# 6 The generalization ability of the neural network

A major disadvantage of previous studies on solving topology optimization problems by deep learning without iteration is that the trained neural network model is limited to work on a specific boundary condition only. Considering the plenty of time needed in the training process, it is impractical to train a new neural network once the boundary conditions change. The proposed method may solve this problem to some extents because the trained neural network in the proposed method has a strong generalization ability. It could work on several different boundary conditions even though all samples in the training dataset are generated from the optimization problem for a cantilever beam boundary. The generalization ability of the trained neural network is demonstrated for two typical engineering boundary conditions (1) a simply supported beam and (2) a two-span continuous beam. The performance is evaluated through 8000 samples for each case. Fig.5 shows some examples of the comparison between the conventional and proposed method. The details of the result are listed in the Table2. Compared with the performance on the cantilever beam boundary condition, the performance on these two boundary conditions is not ideal. The compliance error and the pixel values error are almost doubled and the percentage of the structural disconnection is approximately tripled. Admittedly, changing the boundary condition has some adverse effects on the performance of the neural network. The result still demonstrates the generalization of the neural network considering that it has not 'met' these two kinds of boundary conditions before.

The key factor, which gives the neural network strong generalization ability, may be the format of the neural network's input. In the proposed method, the information about boundary condition is not a direct input for the neural network but hidden in the initial nodal displacements and strains, which are the first five channels of the input tensor. The boundary condition's change is actually the reaction force's change for preparing the input tensor, which is also similar to the force condition's change. Consequently, the neural network can give a less accurate solution to the problem with different boundary condition since each sample has a different force condition in the training set. Nevertheless, the change on boundary condition cannot be completely achieved by the reaction forces, because the rigid displacements would appear when calculating the nodal displacements under the condition that only the external and reaction forces are given. Therefore, the initial nodal displacements must be included in the input tensor. Despite the ingenuity of the input tensor, it is still a challenge for the neural network to predict the optimal structure for a completely new boundary condition. For all the samples in the training set, the displacements of the leftmost nodes are zero, while the displacements of leftmost nodes in the test set are not all zero. It may be a little 'confusing' for the neural network when making the prediction on the test samples and this may explain why the performance on two other boundary conditions is much worse than on the cantilever beam boundary condition. However, the proposed framework is still plausible given the provided generalization ability. In addition, it clearly shows the importance to intelligently design the input to achieve intelligence for the neural network.

# 7 Other discussions about the proposed method

## 7.1 The influence of the sample number on the neural network's performance

It is very expensive to generate the dataset to train the neural network. Too few data may limit the ability and accuracy, while too many data would raise the argument on the benefit. To figure out the relationship between the sample number and neural network's performance and find a trade-off, the neural networks with the same structure were trained by a different number of samples. The numbers of samples include 3000, 4000, 5000, 6000, 7000, 8000, 9000, 10000, 15000, 20000, 25000, 30000, 35000, 40000, 45000, 50000, 55000, 60000, 65000, 70000, 75000, 80000 and samples are divided into training set, validation set and test set with a ratio of 8:1:1 in all experiments. The pixel accuracy on the validation set is shown in the Fig.6. The accuracy at each point was the average result of three same ex-periments. When the training samples number is less than 10000, the pixel accuracy increases obviously with it. From 10000 to 35000, the pixel accuracy increases slightly with the training samples number. After 35000, the pixel accuracy becomes stable (at about 96%). The decline of accuracy with the sample number varying from 7000 to 9000 is a little abnormal. It shows that the neural network's training is unstable if the number of samples is not large enough.

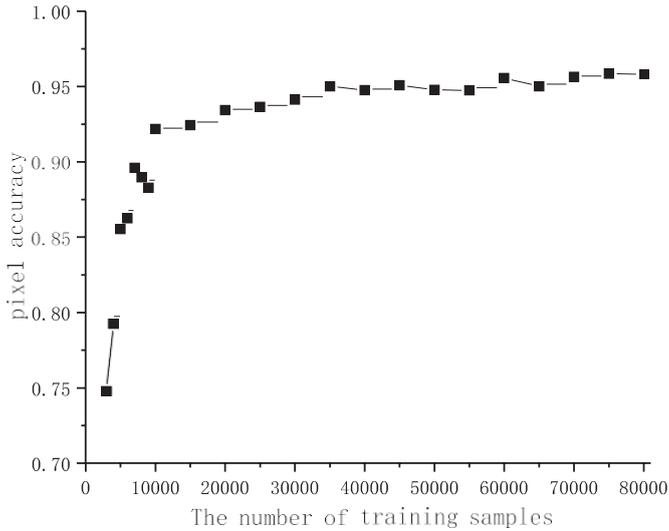

**Fig. 6** The convergence of pixel accuracy on samples number

### 7.2 The influence of input's channels on the neural network's performance

In addition to the final input tensor demonstrated in the section 4.1, two other kinds of input tensors were tried during the experiment. One is a 3-channel tensor only including the volume fraction and nodal displacements in X and Y direction (the first, second and last channels in Fig.1). The other is a 10-channel tensor that includes the all channels in Fig.1 and the other 4 channels proposed in Yu et al. (2018)'s work. The 80000 samples introduced in section 4.1 and 10000 test samples introduced in section 6 were transformed into 3-channel and 10-channel form for training and testing. The same architecture shown in Fig.2 was used to compare the influence of different input tensors on the neural network's performance. Table 3 illustrates the result of average pixel values errors for different input tensors under different boundary conditions.

The performance of 3-channel input tensor is significantly worse than 6-channel input tensor on the cantilever boundary condition. Theoretically, the 3-channel input tensor has all the information in the 6-channel since the strain is the derivative of the displacement, and the neural network should be capable of finding the strain based on the displacement. But the result indicates that an informative input tensor clearly makes it easier for the neural network fit the objective function which can directly give us the near-optimal structure without iteration.

The 6-channel and 10-channel input tensors have almost the same performance on the cantilever boundary condition, which demonstrates that all useful information in the added channels is included in the 6-channel input tensor. Comparing the performance on two other boundary conditions, the average pixel values error is about 25% higher with 6-channel input tensor than 10-channel input tensor. Apparently, the neural network with 6-channel input tensor has better generalization ability and the redundant information in the added channels is adverse to the generalization ability, hence more information in the input tensor is not always better.

**Table 3** Average pixel values errors with different input tensors

| Boundary condition | 3-channel | 6-channel | 10-channel |
|---|---|---|---|
| Cantilever | 13.53% | 4.53% | 4.82% |
| Simply supported | untested | 7.96% | 9.95% |
| Continuous | untested | 7.47% | 9.19% |

*Samples with structural disconnection are not included.

### 7.3 The influence of different architectures on the neural network's performance

Before the final architecture was determined, lots of the architectures had been tested, including GoogLeNet (Szegedy et al., 2014; Ioffe and Szegedy, 2015; Szegedy et al., 2016; Szegedy et al., 2016), ResNet (He et al., 2016) and UNet (Ronneberger et al., 2015). Table 4 shows the performance of different architectures. Each architecture has tried different combinations of hyper-parameters, and the best results are listed in the ta- ble. It turns out that only UNet can significantly im- prove the pixel accuracy comparing to the conventional CNN. Combining the ResNet with the U-Net or mak- ing the neural network deeper cannot improve the accu- racy. The use of GooLeNet (Fig. 7) would increase the training time dramatically but the improvement of performance is almost inconspicuous. Adding the Dropout layer also can not enhance the performance of the neu- ral network, since the the BatchNormalization layers used in the original architecture can almost achieve all Dropout layer's functions. It should be noted that all these conclusions can just provide a reference for other researchers. They are based on the limited hyper-parameters which have been tested in the experiments and it is possible that other researchers find a set of hyper-parameters which can disprove them in the future.

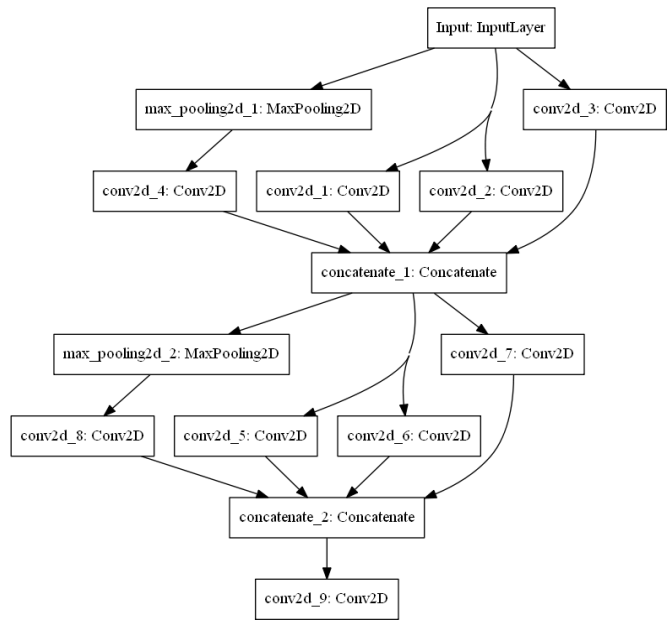

**Fig. 7** The architecture of the GoogLeNet

**Table 4** Performance of different architectures

| The type of architecture | Accuracy on the test set |
| --- | --- |
| Conventional CNN | 81.37% |
| GoogLeNet | 84.95% |
| U-Net | 95.47% |
| U-Net with ResNet | 95.36% |
| U-Net with ResNet and Dropout | 95.42% |

## 8 Conclusion

In this study, a deep CNN is proposed for solving the topology optimization problem. Compared to the conventional method, the proposed method can solve the problem in negligible time. Besides, the strong generalization is the greatest advantage of the proposed method because of the well-designed input form enabling to solve the topology optimization problem with different boundary conditions even though its training datasets have only one boundary condition. This makes it closer to the further practical applications because one well trained deep CNN could be useful in solving all the problems of the same kind, which repays the investment to prepare the datasets and train the neural network.

Nevertheless, the proposed method still has several shortcomings to overcome in future work. The emergence of disconnection, which makes designers can not totally trust the method, is unacceptable for structure final design. Therefore, this method is more suitable to be used in the preliminary design stage when designers want to have a general idea that the well-trained neural network is able to deliver within a very short time. To reduce the disconnection, a new learning objective related to the compliance for the neural network would be required. This will be undertaken as part of our future work. In addition, the input of the neural network must be same size, which means the initial design domain shape cannot change. Besides, the lengthscale problem should be addressed in the future since it is a key issue to obtain a manufacturable structure design. Finally, this paper discusses the multi-load linear elastic compliance minimization problems, which can be solved within a acceptable time range if the scale is small. More work needs to be done in the future to verify the validity of the proposed method in more complicated problems. Although additional efforts are required, the proposed method is still a novel and valuable attempt to use the emerging technique to shed light on accelerating the applications of topology optimization techniques in design practices.

## References

Allaire G, Jouve F, Toader AM (2004) Structural optimization using sensitivity analysis and a level-set method . Journal of Computational Physics 194(1):363–393

Andreassen E, Clausen A, Schevenels M, Lazarov BS, Sigmund O (2011) Efficient topology optimization in matlab using 88 lines of code. Structural Multidisciplinary Optimization 43(1):1–16

Badrinarayanan V, Kendall A, Cipolla R (2017) Seg- net: A deep convolutional encoder-decoder architec- ture for scene segmentation. IEEE Transactions on Pattern Analysis Machine Intelligence PP(99):1–1

Banga S, Gehani H, Bhilare S, Patel S, Kara L (2018) 3d topology optimization using convolu- tional neural networks. CoRR abs/1808.07440, URL http://arxiv.org/abs/1808.07440, 1808.07440

Bendsoe MP (1989) Optimal shape design as a material distribution problem. Structural Optimization 1(4):193–202

Bendsoe MP, Kikuchi N (1988) Generating optimal topologies in structural design using a homogenization method. Computer Methods in Applied Mechanics Engineering 71(2):197–224

Cang R, Yao H, Ren Y (2019) One-shot generation of near-optimal topology through theory-driven machine learning. Computer-Aided Design 109:12 – 21, DOI https://doi.org/10.1016/j.cad.2018.12.008

Du Z, Zhou XY, Picelli R, Kim HA (2018) Connecting microstructures for multiscale topology optimization with connectivity index constraints. Journal of Mechanical Design 140(11):MD–18–1272

Gao G, yu Liu Z, bin Li Y, feng Qiao Y (2017) A new method to generate the ground structure in truss topology optimization. Engineering Optimization 49(2):235–251, DOI 10.1080/0305215X.2016.1169050

Guo T, Lohan DJ, Cang R, Ren MY, Allison JT (2018) An Indirect Design Representation for Topology Optimization Using Variational Autoencoder and Style Transfer. DOI 10.2514/6.2018-0804, URL https://arc.aiaa.org/doi/abs/10.2514/6.2018-0804, https://arc.aiaa.org/doi/pdf/10.2514/6.2018-0804

Guo X, Zhang W, Zhong W (2014) Doing topology optimization explicitly and geometrically—a new moving morphable components based framework. Journal of Applied Mechanics 81(8):081009–081009–12

He K, Zhang X, Ren S, Sun J (2016) Deep residual learning for image recognition. In: IEEE Conference on Computer Vision and Pattern Recognition, pp 770–778

Ioffe S, Szegedy C (2015) Batch normalization: accelerating deep network training by reducing internal covariate shift. In: International Conference on International Conference on Machine Learning, pp 448–456

Kingma DP, Ba J (2014) Adam: A Method for Stochastic Optimization. ArXiv e-prints arXiv:1412.6980, 1412.6980

Lecun Y, Bottou L, Bengio Y, Haffner P (1998) Gradient-based learning applied to document recognition. Proceedings of the IEEE 86(11):2278–2324

Lei X, Liu C, Du Z, Zhang W, Guo X (2018) Machine learning driven real time topology optimization under moving morphable component (mmc)-based framework. Journal of Applied Mechanics 86(1):011004–9

Michell A (1904) Lviii. the limits of economy of material in frame-structures. Philosophical Magazine 8(47):589–597

Nagarajan HPN, Mokhtarian H, Jafarian H, Di- massi S, Bakrani-Balani S, Hamedi A, Coatanéa E, Gary Wang G, Haapala KR (2018) Knowledge-based design of artificial neural network topology for additive manufacturing process modeling: A new approach and case study for fused deposition modeling. Journal of Mechanical Design 141(2):MD–18–1545

Nair V, Hinton GE (2010) Rectified linear units improve restricted boltzmann machines. In: International Conference on International Conference on Machine Learning, pp 807–814

Oh S, Jung Y, Kim S, Lee I, Kang N (2019) Deep Generative Design: Integration of Topology Optimization and Generative Models. Journal of Mechanical Design 141(11), DOI 10.1115/1.4044229, URL https://doi.org/10.1115/1.4044229, 111405

Rawat S, Shen MHH (2018) A novel topology design approach using an integrated deep learning network architecture

Ronneberger O, Fischer P, Brox T (2015) U-net: Convolutional networks for biomedical image segmentation. In: International Conference on Medical Image Computing and Computer-Assisted Intervention, pp 234–241

Rozvany GIN, Zhou M, Birker T (1992) Generalized shape optimization without homogenization. Structural Optimization 4(3-4):250–252

Shelhamer E, Long J, Darrell T (2014) Fully convolutional networks for semantic segmentation. IEEE Transactions on Pattern Analysis Machine Intelligence 39(4):640–651

Sigmund O, Maute K (2013) Topology optimization approaches. Structural Multidisciplinary Optimization 48(6):1031–1055

Sosnovik I, Oseledets I (2017) Neural networks for topology optimization. ArXiv e-prints 1709.09578

Szegedy C, Liu W, Jia Y, Sermanet P, Reed S, Anguelov D, Erhan D, Vanhoucke V, Rabinovich A (2014) Going Deeper with Convolutions. ArXiv e-prints 1409.4842

Szegedy C, Ioffe S, Vanhoucke V, Alemi A (2016) Inception-v4, Inception-ResNet and the Impact of Residual Connections on Learning. ArXiv e-prints 1602.07261

Szegedy C, Vanhoucke V, Ioffe S, Shlens J, Wojna Z (2016) Rethinking the inception architecture for computer vision. In: Computer Vision and Pattern Recognition, pp 2818–2826

Wang MY, Wang X, Guo D (2003) A level set method for structural topology optimization. Computer Methods in Applied Mechanics and Engineering 192(1–2):227–246

Xie YM, Steven GP (1993) A simple evolutionary procedure for structural optimization. Computers Structures 49(5):885–896

Yang Z, Li X, Catherine Brinson L, Choudhary AN, Chen W, Agrawal A (2018) Microstructural materials design via deep adversarial learning methodology. Journal of Mechanical Design 140(11):MD–18–1252

Yu Y, Hur T, Jung J, Jang IG (2018) Deep learning for determining a near-optimal topological design without any iteration. Structural and Multidisciplinary Optimization DOI 10.1007/s00158-018-2101-5, URL https://doi.org/10.1007/s00158-018-2101-5

Zhang W, Liu Y, Du Z, Zhu Y, Guo X (2018) A moving morphable component based topology optimization approach for rib-stiffened structures considering buckling constraints. Journal of Mechanical Design 140(11):MD–18–1182

Zhou M, Rozvany G (1991) The coc algorithm, part ii: Topological, geometrical and generalized shape optimization. Computer Methods in Applied Mechanics and Engineering 89(1):309 – 336, DOI https://doi.org/10.1016/0045-7825(91)90046-9, second World Congress on Computational Mechanics